\documentclass{article}

\usepackage[nonatbib,final]{nips_2017}
\usepackage[colorlinks,
linkcolor=black,
anchorcolor=black,
citecolor=black,
urlcolor=black
]{hyperref}
\usepackage[utf8]{inputenc} % allow utf-8 input
\usepackage[T1]{fontenc}    % use 8-bit T1 fonts
\usepackage{booktabs}       % professional-quality tables
\usepackage{amsfonts}       % blackboard math symbols
\usepackage{nicefrac}       % compact symbols for 1/2, etc.
\usepackage{microtype}      % microtypography

\usepackage[numbers]{natbib}
\usepackage{graphicx}
\usepackage{algorithm}
\usepackage{algorithmic}

\usepackage{amssymb}
\usepackage{amsthm}
\usepackage{amsmath}
\usepackage{multirow}
\usepackage{mathrsfs}
\usepackage{bbm}
\usepackage{bm}
\usepackage{eufrak}
\usepackage{bookmark}
\usepackage{mathtools}

\graphicspath{{illustrations/}}

\title{Learning $3$D-FilterMap for Deep Convolutional Neural Networks}

\author{
Yingzhen Yang$^{1}$, Jianchao Yang$^{1}$, Ning Xu$^{1}$, Wei Han$^{2}$\\
$^{1}$ Snap Research, Venice, CA 90291, USA
$^{2}$ Beckman Institute for Advanced Science and Technology,\\
University of Illinois at Urbana-Champaign, Urbana, IL 61801, USA \\
\texttt{superyyzg@gmail.com, \{jianchao.yang,ning.xu\}@snap.com, weihan3@illinois.edu} \\
}

\newcommand{\bF}{\mathbf{F}}
\newcommand{\bG}{\mathbf{G}}

\newcommand{\bT}{\mathbf{T}}

\newcommand{\N}{{\rm I}\kern-0.18em{\rm N}}

\newcommand{\R}{{\rm I}\kern-0.18em{\rm R}}
\newcommand{\h}{{\rm I}\kern-0.18em{\rm H}}
\newcommand{\K}{{\rm I}\kern-0.18em{\rm K}}
\newcommand{\p}{{\rm I}\kern-0.18em{\rm P}}
\newcommand{\E}{{\rm I}\kern-0.18em{\rm E}}
\newcommand{\Z}{{\rm Z}\kern-0.18em{\rm Z}}

\newcommand{\1}{{\rm 1}\kern-0.25em{\rm I}}

\newcommand{\pn}{\p_{\kern-0.25em n}}
\newcommand{\pnm}{\p_{\kern-0.25em n,m}}
\newcommand{\psubm}{\p_{\kern-0.25em m}}

\newcommand{\BigO}[1]{{\operatorname{O}}}

\newtheorem*{MyTheorem*}{Theorem}
\newtheorem*{MyLemma*}{Lemma}
\begin{document}

\maketitle

\begin{abstract}
  We present a novel and compact architecture for deep Convolutional Neural Networks (CNNs) in this paper, termed $3$D-FilterMap Convolutional Neural Networks ($3$D-FM-CNNs). The convolution layer of $3$D-FM-CNN learns a compact representation of the filters, named $3$D-FilterMap, instead of a set of independent filters in the conventional convolution layer. The filters are extracted from the $3$D-FilterMap as overlapping $3$D submatrics with weight sharing among nearby filters, and these filters are convolved with the input to generate the output of the convolution layer for $3$D-FM-CNN. Due to the weight sharing scheme, the parameter size of the $3$D-FilterMap is much smaller than that of the filters to be learned in the conventional convolution layer when $3$D-FilterMap generates the same number of filters. Our work is fundamentally different from the network compression literature that reduces the size of a learned large network in the sense that a small network is directly learned from scratch. Experimental results demonstrate that $3$D-FM-CNN enjoys a small parameter space by learning compact $3$D-FilterMaps, while achieving performance compared to that of the baseline CNNs which learn the same number of filters as that generated by the corresponding $3$D-FilterMap.
\end{abstract}

\section{Introduction}
With the rise of deep learning, deep Convolutional Neural Networks (CNNs) are popular choices for learning highly semantic and discriminative features for various tasks including image classification. CNN belongs to the feedforward network, and it incorporates a sequence of convolution processes with the key idea of local receptive field, shared weights among the neurons and subsampling processes which reduces the spatial resolution of the activation map typically by max-pooling \cite{JarrettKRL09}. In each convolutional layer of CNN, the input volume is convolved with a set of separate filters in that layer, producing the activation map. The filters are learned in a way such that they incur strong response for certain types of feature at some spatial position of the input. The subsampling process by max-pooling then takes the maximum response within local windows of the input volume (which is always the output of the convolutional layer), achieving a certain degree of invariance to deformations in the input.

It can be observed that the convolutional layer is important for learning deformation invariant features, e.g. translation invariant features, by convolving the input with filters, and the number of filters influences the quantity of such features. On the other hand, it is widely believed that there is considerable redundancy in the filters, and previous study shows that the parameter space of the filters can be significantly reduced by various model compression techniques such as quantization or exploiting the low-rank and sparse representation of the filters \cite{HanMD15-deep-compression,Ioannou2016-low-rank-filter,Yu2017-low-rank-sparse-compression}

Based on the observation of the redundancy in the filters, the above discussion leads us to an interesting question: is there a way of generating the filters from a reduced parameter space? If it is achievable, we then have a solution that learns a compact CNN from scratch. We propose $3$D-FilterMap as a novel and compact structure of organizing the filters to achieve this goal. $3$D-FilterMap is a $3$D matrix from which the filters are extracted from as overlapping $3$D submatrices (see Figure~\ref{fig:3d-filtermap}). When a certain number of filters are extracted from a $3$D-FilterMap wherein nearby filters share weights, the parameter space of the $3$D-FilterMap is much smaller than that of the same number of independent filters to be learned in the conventional convolution layer. In this manner, $3$D-FilterMap enables a way of directly learning a compact CNN. In contrast, the model compression literature broadly adopts a two-step approach: learning a large CNN first, then compressing the model by various model compression techniques such as pruning, quantization and coding \cite{HanMD15-deep-compression,LuoWL17-ThiNet}, or utilizing the low-rank or sparse representation of the filters based on the redundancy in them \cite{Ioannou2016-low-rank-filter,Yu2017-low-rank-sparse-compression}.

The detailed formulation of $3$D-FilterMap and the associated convolutional network that employs $3$D-FilterMap are introduced in the next section.

\section{Formulation}
The idea of $3$D-FilterMap is inspired by epitome \cite{JojicFK03}, which is developed in the computer vision and machine learning literature for learning a condensed version of Gaussian Mixture Models (GMMs). In epitome, the Gaussian means are represented by a two dimensional matrix wherein each window in this matrix contains parameters of the Gaussian means for a Gaussian component. The same structure is adopted for representing the Gaussian covariances. If the number of non-overlapping windows in the mean matrix and the covariances matrix is the same as the number of Gaussian components in the conventional GMMs, the epitome possesses significantly more number of Gaussian components than its GMMs counterpart since much more Gaussian means and covariances can be extracted densely from the mean and covariances matrices of the epitome. Therefore, the generalization and representation capability of epitome outshines GMMs with the same parameter space, while circumventing the potential overfitting.

The above characteristics of epitome encourages us to arrange filters in a way similar to epitome in the proposed $3$D-FilterMap Convolutional Neural Networks ($3$D-FM-CNN). More concretely, each convolution layer of $3$D-FM-CNN has a $3$D matrix named $3$D-FilterMap, wherein the overlapping $3$D submatrices in the $3$D-FilterMap play the same role as the filters in the conventional convolution layer of ordinary CNN. $3$D-FM-CNN and its baseline CNN have the same architecture except that each convolution layer of $3$D-FM-CNN comprises a $3$D-FilterMap rather than a set of independent filters. For each convolution layer of $3$D-FM-CNN, the $3$D-FilterMap is designed to have a proper size such that a certain number of filters can be generated by densely extracting overlapping $3$D submatrices from it, while the redundancy of the filters is removed by sharing weights across nearby filters in their overlapping region. The $3$D-FilterMap in each convolution layer of $3$D-FM-CNN generates the same number of filters as the corresponding convolution layer in the baseline CNN. Instead of learning a set of independent filters for each convolution layer of CNN, a compact $3$D-FilterMap of much smaller parameter size is learned for each layer of $3$D-FM-CNN.

\begin{figure*}[!htb]
\begin{center}
\includegraphics[width=0.23\textwidth]{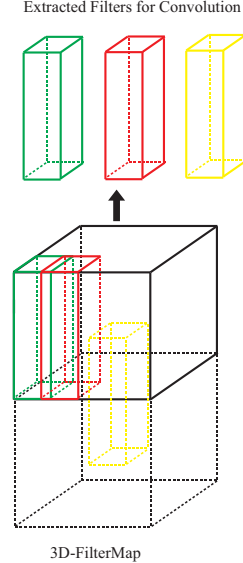}
\end{center}
   \caption{Illustration of a $3$D-FilterMap. The two filters in green and red are two overlapping filters extracted from the $3$D-FilterMap. A copy of $3$D-FilterMap, which is illustrated by dashed lines, is attached along the channel dimension to the original $3$D-FilterMap so that filters such as the one in yellow can be extracted using ``wraparound'' weights. A $3$D-FilterMap of size $64 \times 8 \times 8$ is learned in the convolution layer of $3$D-FM-CNN in the case of $64$ $64 \times 3 \times 3$ filters in the corresponding convolution layer of the baseline CNN. }
\label{fig:3d-filtermap}
\end{figure*}

An illustration of $3$D-FilterMap is shown in Figure~\ref{fig:3d-filtermap} which illustrates an example of $3$D-FilterMap and how overlapping filters are extracted from it. Suppose that a convolution layer of the baseline CNN model has $64$ filters of channel $64$ and spatial size $3 \times 3$, the corresponding convolution layer in the $3$D-FM-CNN has a $3$D-FilterMap of size $64 \times 8 \times 8$. The $64$ $64 \times 3 \times 3$ filters are sampled by striding along each spatial dimension by $2$, and striding along the dimension of the channel by $16$. The ratio of the parameter size of the $64$ independent filters to that of the corresponding $3$D-FilterMap is $\frac{64 \times 64 \times 3 \times 3}{64 \times 8 \times 8} = 9$, indicating that the parameter space of the $3$D-FilterMap is much smaller than the independent filters in the baseline CNN.

Formally, suppose a $3$D-FilterMap should generate $K = K_1 \times K_2 \times K_3$ filters of size $S_1 \times S_2 \times C$ where $(S_1,S_2)$ is the spatial size of filter and $C$ is the channel size. Let the filter sampling stride along two spatial dimensions of the $3$D-FilterMap are $x$ and $y$, and the sampling stride along the channel dimension of the $3$D-FilterMap is $z$. Then the dimension of the $3$D-FilterMap is $(K_1x,K_2y,K_3z)$, where $(K_1x,K_2y)$ is the spatial size and $K_3z$ is the channel size. In this paper we set the channel size of the $3$D-FilterMap to be $K_3z = C$, which is based on our observation that the weights along the channel can be shared more frequently without hurting the performance. Therefore, the ratio of the parameter size of the $K$ independent filters to that of the corresponding $3$D-FilterMap is

\begin{align*}
&{\rm ParamRatio} = \frac{K \cdot S_1 \cdot S_2 \cdot C}{K_1x \cdot K_2y \cdot K_3z} = \frac{K_1K_2K_3 \cdot S_1 \cdot S_2 \cdot C}{K_1x \cdot K_2y \cdot K_3z} = \frac{S_1 \cdot S_2 \cdot C}{x \cdot y \cdot \cdot z} = \frac{S_1 \cdot S_2}{x \cdot y \cdot} \cdot K_3
\end{align*}

In a typical setting where the spatial stride is smaller than the corresponding filter size, i.e. $x < S_1$, $y < S_2$, the $3$D-FilterMap has a compact size. Also note that a larger $K_3$, namely the sampling number along the channel dimension, leads to a more compact $3$D-FilterMap in the manner that the weights of the $3$D-FilterMap along the channel dimension are shared more frequently by nearby filters.

Algorithm~\ref{alg:3D-filtermap-forward-backward} describes the forward and backward operation in a convolution layer of $3$D-FM-CNN with $3$D-FilterMap. We use the mapping $\bT$ which maps the indices of the elements of the extracted filters to the indices of the corresponding element in the $3$D-FilterMap. Namely, for a filter $\bF^{(k)}$ and the $3$D-FilterMap $\bF^{(M)}$, $\bF_t^{(k)} = \bF_{\bT(t)}^{(M)}$ (please refer to the notations in Algorithm~\ref{alg:3D-filtermap-forward-backward}). The mapping $\bT$ is used to conveniently track the origin of the elements of the filters extracted from the $3$D-FilterMap.

\begin{algorithm}[!ht]
\renewcommand{\algorithmicrequire}{\textbf{Input:}}
\renewcommand\algorithmicensure {\textbf{Output:} }
\caption{Description of Forward and Backward Operation in a convolution layer of $3$D-FM-CNN with $3$D-FilterMap}
\label{alg:3D-filtermap-forward-backward}
%\allowdisplaybreaks
\begin{algorithmic}[1]
\REQUIRE ~~\\
%$\lambda_{\ell^{1}}$ for the initialization of the the A$\ell^{0}$-SSC,
%maximum iteration number $M$, stopping threshold $\varepsilon$\\
\STATE \textbf{Forward:} Extract $K$ overlapping filters $\{\bF^{(k)}\}_{k=1}^K$ from the $3$D-FilterMap $\bF^{(M)}$, and each filter $\bF^{(k)} \in \R^{S_1 \times S_2 \times C}$. Then convolve these $K$ filters with the input of the convolution layer, where $K$ is the number of filters in the corresponding convolution layer of the baseline CNN.
\STATE \textbf{Backward:} First obtain the gradient of all the $K$ filters as $\{\bG^{(k)}\}_{k=1}^K$ where each $\bG^{(k)} \in \R^{S_1 \times S_2 \times C}$. For each element $j \in \bF^{(M)}$, its gradient is computed by
\begin{align}\label{eq:gradient-j}
&j' = \frac{ \sum\limits_{k=1}^K \sum\limits_t \bG_t^{(k)} \1_{\bT(t)=j} }  { \sum\limits_{k=1}^K \sum\limits_t \1_{\bT(t)=j} }
\end{align}
%\WHILE{$t \le M$}
%\STATE{Obtain ${\balpha}^{(t)}$ from ${\balpha}^{(t-1)}$ by (\ref{eq:l0ssc-lasso-proximal-step1}) and (\ref{eq:l0ssc-lasso-proximal-step2})}
%\IF{$|L(\balpha^{(t)})-L(\balpha^{(t-1)})| < \varepsilon$}
%%\PRINT
%\STATE \textbf{break}
%\ELSE
%\STATE{$t=t+1$.}
%\ENDIF
%%and in each step of coordinate descent use the feature-sign search algorithm to solve the optimization problem (\ref{eq:objfunci}).
%\ENDWHILE
\end{algorithmic}
\end{algorithm}

%In addition to capability of model compression, $3$D-FilterMap inherently enables a better visualization of the filters by forming a “panorama” of the filters such that adjacent filters changes smoothly across the spatial domain, and similar filters group together. A visualization of the learned filter map is illustrated in Figure 2 (note that this is only for demonstration purpose; a much smaller filter map is used for model compression). This feature would benefit the design of the networks and make it easier to observe the characteristics of the filters learned in the convolutional neural networks. Figure 1 shows the illustration of $3$D-FM-CNN.

\section{Experimental Results}
We demonstrate the performance of $3$D-FM-CNN in this section by the comparative results between $3$D-FM-CNN and its baseline CNNs. Using ResNet \cite{He2016-resnet} and DenseNet \cite{HuangLW16a-densenet} as the baseline CNNs, we design $3$D-FM-CNN by replacing each $3 \times 3$ convolution layer of ResNet or DenseNet by a convolution layer with $3$D-FilterMap in the $3$D-FM-CNN. We train $3$D-FM-CNN and the baseline CNNs on the CIFAR-10 data, and show the testing accuracy and the parameter number of all the models in Table~\ref{table:3D-FM-CNN}. ResNet-18-FM indicates the $3$D-FM-CNN using ResNet-18 as the baseline CNN, and similar naming convention is used for other models. The convolution layers of ResNet and DenseNet have filters of spatial size of either $3 \times 3$ or $1 \times 1$. We replace all the $3 \times 3$ convolution layers by the convolution layers with $3$D-FilterMap in the corresponding $3$D-FM-CNN so as to demonstrate the capability of $3$D-FilterMap to represent the filters that capture the spatial pattern in the input. Note that we do not use $3$D-FilterMap to represent $1 \times 1$ convolution layers since $1 \times 1$ convolution layer is primarily used for dimension reduction. We design the size of the $3$D-FilterMap according to the number of filters in the corresponding convolution layer of the baseline CNN. Throughout this section, we set the size of the $3$D-FilterMap according to Table~\ref{table:fm-size}, and use the spatial stride $x = y = 2$. We do not specifically tune $(K_1,K_2,K_3)$, and one can surely choose other settings for these three hyperparameters as long as their product matches the number of filters in the baseline CNN.

It can be observed in Table~\ref{table:3D-FM-CNN} that $3$D-FM-CNN achieves comparable performance with a compact parameter space for different baseline CNNs including ResNet-$18$, ResNet-$34$, ResNet-$50$ and DenseNet-$121$. DenseNet-$121$ indicates DenseNet with a growth rate of $32$ and $121$ layers. Since most of the parameters of ResNet-$18$ and ResNet-$34$ are in the $3 \times 3$ convolution layers, 3D-FM-CNN exhibits a relatively high parameter ratio (ratio of the parameter number of the baseline CNN to that of the 3D-FM-CNN) of around $11.7$. ResNet-$50$ and DenseNet-$121$ have a significant amount of parameters in $1 \times 1$ convolution layers, therefore, the parameter ratio is not that high. However, in the latter case $3$D-FM-CNN slightly generalizes better evidenced by an even better testing accuracy.

\begin{table*}[ht]
\centering
\scriptsize
\caption{\small The Size of $3$D-FM-CNN}
\begin{tabular}{|c|c|c|c|c|}
  \hline

  \#Filters      &$K_1$  &$K_2$  &$K_3$  \\ \hline

  12             &2      &3      &2    \\ \hline
  32             &4      &4      &2    \\ \hline
  64             &4      &4      &4    \\ \hline
 128             &8      &4      &4    \\ \hline
 256             &8      &8      &4    \\ \hline
 512             &8      &8      &8    \\ \hline
\end{tabular}
\label{table:fm-size}
\end{table*}
\begin{table*}[!ht]
\centering
\scriptsize
\caption{\small Performance of $3$D-FM-CNN and the corresponding baseline CNN}
\begin{tabular}{|c|c|c|c|c|c|c|c|c|c|c|}
  \hline

  Model    &ResNet-$18$    &ResNet-$18$-FM  &ResNet-$34$  &ResNet-$34$-FM   &ResNet-$50$  &ResNet-$50$-FM   &DenseNet-$121$  &DenseNet-$121$-FM \\ \hline

Accuracy    &94.18\%    &93.55\%       &94.72\%    &94.25\%        &95.16\%    &95.47\%        &95.13\%      &95.45\%        \\ \hline
\#Parameter &11.2M      &0.95M         &21.3M      &1.8M           &23.5M      &13.1M          &7.0M       &5.3M         \\ \hline

\end{tabular}
\label{table:3D-FM-CNN}
\end{table*}

\section{Conclution}
We present $3$D-FilterMap Convolutional Neural Networks ($3$D-FM-CNNs) in this paper. In contrast with learning a set of independent filters in the conventional convolution layer, the convolution layer of $3$D-FM-CNN learns a compact $3$D-FilterMap. The acutal filters are extracted from the $3$D-FilterMap in a way such that nearby filters share weights. Thanks to the weight sharing scheme,
$3$D-FM-CNN has a much smaller parameter space than its baseline CNN when it generates the same number of filters as the baseline CNN. Experimental results demonstrate the effectiveness of $3$D-FM-CNN in learning a compact model while exhibiting performance comparable to the baseline CNNs.
\small{
\bibliographystyle{unsrt}
\bibliography{mybib}

\begin{thebibliography}{1}

\bibitem{JarrettKRL09}
Kevin Jarrett, Koray Kavukcuoglu, Marc'Aurelio Ranzato, and Yann LeCun.
\newblock What is the best multi-stage architecture for object recognition?
\newblock In {\em {IEEE} 12th International Conference on Computer Vision,
  {ICCV} 2009, Kyoto, Japan, September 27 - October 4, 2009}, pages 2146--2153,
  2009.

\bibitem{HanMD15-deep-compression}
Song Han, Huizi Mao, and William~J. Dally.
\newblock Deep compression: Compressing deep neural network with pruning,
  trained quantization and huffman coding.
\newblock In {\em Proceedings of the International Conference on Learning
  Representations (ICLR)}, 2015.

\bibitem{Ioannou2016-low-rank-filter}
Yani Ioannou, Duncan~P. Robertson, Jamie Shotton, Roberto Cipolla, and Antonio
  Criminisi.
\newblock Training cnns with low-rank filters for efficient image
  classification.
\newblock In {\em Proceedings of the International Conference on Learning
  Representations (ICLR)}, 2016.

\bibitem{Yu2017-low-rank-sparse-compression}
Xiyu Yu, Tongliang Liu, Xinchao Wang, and Dacheng Tao.
\newblock On compressing deep models by low rank and sparse decomposition.
\newblock In {\em 2017 {IEEE} Conference on Computer Vision and Pattern
  Recognition, {CVPR} 2017, Honolulu, HI, USA, July 21-26, 2017}, 2017.

\bibitem{LuoWL17-ThiNet}
Jian{-}Hao Luo, Jianxin Wu, and Weiyao Lin.
\newblock Thinet: {A} filter level pruning method for deep neural network
  compression.
\newblock In {\em 2017 {IEEE} International Conference on Computer Vision,
  {ICCV} 2017, Venice, Italy, October 22-29, 2017}, 2017.

\bibitem{JojicFK03}
Nebojsa Jojic, Brendan~J. Frey, and Anitha Kannan.
\newblock Epitomic analysis of appearance and shape.
\newblock In {\em 9th {IEEE} International Conference on Computer Vision
  {(ICCV} 2003), 14-17 October 2003, Nice, France}, pages 34--43, 2003.

\bibitem{He2016-resnet}
K.~He, X.~Zhang, S.~Ren, and J.~Sun.
\newblock Deep residual learning for image recognition.
\newblock In {\em 2016 IEEE Conference on Computer Vision and Pattern
  Recognition (CVPR)}, pages 770--778, June 2016.

\bibitem{HuangLW16a-densenet}
Gao Huang, Zhuang Liu, and Kilian~Q. Weinberger.
\newblock Densely connected convolutional networks.
\newblock In {\em 2017 {IEEE} Conference on Computer Vision and Pattern
  Recognition, {CVPR} 2017, Honolulu, HI, USA, July 21-26, 2017}, 2017.

\end{thebibliography}

}

\end{document}